\documentclass[runningheads]{llncs}
\usepackage{amsfonts,amssymb}
\usepackage{graphicx}
\usepackage{booktabs,multirow}
\usepackage{amssymb}
\usepackage{amsmath}
\usepackage{amstext}
\usepackage{comment}
\usepackage{color}
\usepackage{pifont}
\usepackage{bbding}
\usepackage{amsfonts,amssymb,amsmath}

\usepackage[pagebackref=true,breaklinks=true,colorlinks=true,bookmarks=false,citecolor=blue,urlcolor=black,linkcolor=blue]{hyperref}

\usepackage{todonotes}

\usepackage{amsmath}

\usepackage[marginal]{footmisc}

\begin{document}
%

\title{Robotic Surgery Remote Mentoring via AR with 3D Scene Streaming and Hand Interaction}
\titlerunning{AR-based Remote Mentoring for Robotic Surgery}
%
\author{Yonghao Long* \and
Chengkun Li* \and
Qi Dou}
\institute{Dept. of Computer Science and Engineering, The Chinese University of Hong Kong}

%

%
%
\maketitle              
\vspace{-4mm}
\begin{abstract}
With the growing popularity of robotic surgery, education becomes increasingly important and urgently needed for the sake of patient safety.
However, experienced surgeons have limited accessibility due to their busy clinical schedule or working in a distant city, thus can hardly provide sufficient education resources for novices.
Remote mentoring, as an effective way, can help solve this problem, but traditional methods are limited to plain text, audio, or 2D video, which are not intuitive nor vivid. 
Augmented reality (AR), a thriving technique being widely used for various education scenarios, is promising to offer new possibilities of visual experience and interactive teaching.
In this paper, we propose a novel AR-based robotic surgery remote mentoring system with efficient 3D scene visualization and natural 3D hand interaction. Using a head-mounted display (i.e., HoloLens), the mentor can remotely monitor the procedure streamed from the trainee's operation side. The mentor can also provide feedback directly with hand gestures, which is in-turn transmitted to the trainee and viewed in surgical console as guidance.
We comprehensively validate the system on both real surgery stereo videos and ex-vivo scenarios of common robotic training tasks (i.e., peg-transfer and suturing). Promising results are demonstrated regarding the fidelity of streamed scene visualization, the accuracy of feedback with hand interaction, and the low-latency of each component in the entire remote mentoring system.
This work showcases the feasibility of leveraging AR technology for reliable, flexible and low-cost solutions to robotic surgical education, and holds great potential for clinical applications.
\vspace{-2mm}
\vspace{-7mm}
\footnote{\noindent *Authors contributed equally to this work.}

\end{abstract}
\section{Introduction}
\vspace{-2mm}

Robotic surgery has been revolutionized in past decades~\cite{navab2021robotic,d2021accelerating}, greatly enhancing the surgeon's perception and operational ability in minimally invasive interventions. With dramatic rise of adoption of robotic surgery, education is important and requires urgent enhancement~\cite{chen2021evolving} for teaching the next generation of surgeons how to use robots to conduct a surgery step-by-step, instead of using traditional laparoscopic paradigms. Unfortunately, the novices always have limited access to experienced surgeons who have intensive clinical schedule or work in a distant city, making the surgical education currently challenging and expensive~\cite{collins2021telementoring,erridge2019telementoring,huang2019telemedicine}. 


Remote mentoring, i.e., experts monitor surgical progress and provide guidance for trainees from a different geographical location, is envisaged as a promising way to reduce cost and facilitate surgical education. 
Hinata et al.~\cite{hinata2014novel} employ video-audio communication and transfer handwritten notes with a touch tablet for remote mentoring. Their results present similar surgical outcomes (e.g., operating time, complication, early continence status, positive margin rate) between the remote and onsite mentoring groups. 
Mendez et al.~\cite{mendez2005robotic} propose to transmit video, audio communication and surgical instructions with electronic stylus and pad, for a $400km$ long-distance mentoring in neurosurgery. Experiments validate its safety and trustworthiness.
However, common shortcomings of these methods are that, the mentoring and monitoring information are conveyed through text, voice, image and 2D video, which are not intuitive nor vivid for perceiving rich circumstances of surgery and provide comprehensive feedback by mentors.

Recently, augmented reality (AR) brings new possibility to remote mentoring along with many benefits~\cite{yoon2018augmented}.
AR can provide an immersive way of displaying the 3D information (e.g., surgical scene and tool movement) and allow easier and natural interactions~\cite{barmaki2019enhancement,kovoor2021validity,qian2019review}.
Some initial investigations exist with encouraging progress.
For examples, Jarc et al.~\cite{jarc2017proctors} overlay ghost tools for guidance within the view of trainee's console; the mentor sees the surgical scene with a 3D display and glass, and controls the ghost tools through an extra handle device.
Shabir et al.~\cite{shabir2021towards} propose a robotic surgery remote mentoring system, which uses haptic devices to control virtual tools, but with a high latency observed even in a local area network.
Though much benefits have been demonstrated, existing solutions still cannot achieve high-fidelity streaming of 3D scenes via AR, which constrains both efficiency and quality of remote mentoring. In addition, the requirement of extra interaction devices would unavoidably lead to inconvenience of education.

In this paper, to address current limitations, we design a novel robotic surgery remote mentoring system relying on AR with 3D surgical scene streaming and hand interaction. 
Specifically, we first design an efficient streaming scheme to transmit the 3D surgical scene from the operation side to mentor side, using color and disparity images to reduce transmission cost. Then, we allow the mentor to use a portable head-mounted display to perceive 3D surgical scene in point cloud representation and give feedback guidance with hand gestures conveniently. Last, the instructions are returned to the trainee side and can be viewed in the console for straight-forward guidance.
The system is validated on both real surgery video data and common ex-vivo scenarios of robotic training tasks, demonstrating promising results in terms of fidelity of streamed scene visualization, accuracy of hand interaction, and low-latency of transmission.
To the best of our knowledge, this is the first completely AR-based remote mentoring system with only a head-mounted display for robotic surgical education.
This new type of reliable, flexible and low-cost solution has great potential to be used in clinical practice.

\vspace{-2mm}
\section{Methods and Materials}
\vspace{-2mm}


\begin{figure}[t]
\includegraphics[width=1\textwidth]{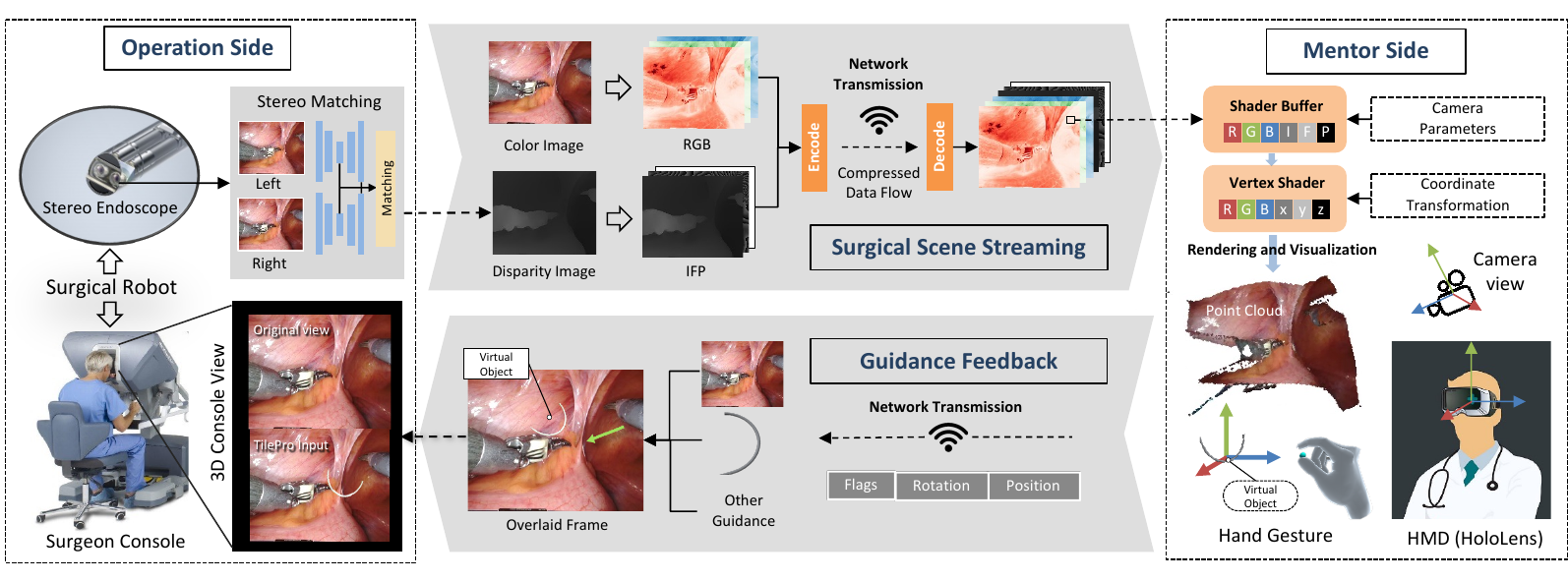}
\vspace{-9mm}
\caption{The overview of our proposed novel system for robotic surgery remote mentoring via augmented reality (AR) with 3D scene streaming and hand interaction in HoloLens.}
\label{Fig_pipeline}
\vspace{-6mm}
\end{figure}

The overview of our AR-based robotic surgery remote mentoring system is shown in Fig.~\ref{Fig_pipeline}. It consists of four components: operation side, surgical scene streaming, mentor side with hand interaction, guidance feedback for closed-loop education.

Briefly speaking, in the operation side, a trainee conducts training tasks or a surgery at the console, which shows stereo surgical video, as well as the mentor's guidance information displayed in AR via TilePro toolkit (Intuitive Surgical Inc).
For surgical scene streaming, the RGB-D information of 3D scene is efficiently transmitted via high-speed internet from the operation side to the remote mentor side. Then, the mentor (i.e., an expert surgeon) uses a head-mounted AR glass to see the surgical scene in real-time and accordingly provide guidance with hand gestures. 
Finally, the mentor's inputs with virtual objects are streamed back and visualized in trainee's console as a closed-loop education system.

\vspace{-2mm}
\subsection{Efficient Surgical Scene Representation and Streaming}
\vspace{-2mm}
To represent the surgical scene, we opt for RGB-D as 3D representation in operation side, owing to the advantages that it can accurately describe relatively complex objects with finite points. It can also be easily obtained without heavy reconstruction computation compared to mesh, as well as can be straightforwardly and efficiently projected to point clouds for 3D visualization. Here, we first apply the HITNet \cite{tankovich2021hitnet}, a real-time deep learning method which can extract the high-level multi-scale context features from a stereo image pair $(I_L, I_R)$ to calculate the correspondence for each pixel, and yield the disparity map $\boldsymbol{D}$ of the left image. 
These RGB and disparity images are leveraged for transmission, in order to save memory and keep up the streaming speed for fast system responses.

Different from previous work, we design a new compression operation to reduce the burden of wireless communication in-between mentor-operation sides. Qian et al.~\cite{qian2019aramis} suggests to transmit flattened RGB-D with three-bytes RGB color and one-byte disparity value. However, it would sacrifice quality due to compact representation which compresses a float type disparity value to one-byte type. Apart from this, the flatten array representation will also cause heavy network communication pressure as the frame resolution increasing. For instance, in the case of $640 \times 512$ resolution, it would cost $640 \times 512 \times (3 + 1) \times 8$ bits $= 10$ Mb for every frame. 
Instead, our calculated disparity map (which is a float single channel image) is separated to two channels, with one denoting the integer part ($I$ channel) of the disparity map, and the other for the fractional part ($F$ channel). These two channels are then combined with an all-zero placeholder $P$ channel to generate a 3-channel image to fulfill the format requirement of the efficient ``JPEG" compression algorithm, whose information loss and computational time will not exhibit exponential growth with larger data.  Finally, the IFP image and corresponding RGB frame are both encoded as ``JPEG" format and then combined to be our surgical scene streaming data flow. In this way, we can reduce the transmission data size to around 8 times smaller than previous work~\cite{qian2019aramis}, i.e. 1.25 Mb for the resolution of $640 \! \times \! 512$.
The data flow is then sent to mentor side using the internet socket TCP/IP. 
Similarly, we do ``JPEG" decoding in remote side to obtain RGB and IFP images for further processing.

\vspace{-2mm}
\subsection{HMD Visualization and Hand-based Natural Interaction}
\vspace{-2mm}
As mentor side is an expert who requires to setup and use the system as convenient as possible, we leverage optical see-through HMD to fulfill the requirement. Specifically, we fist convert the IFP channels to disparity image and transfer it to a point cloud with the following projection equation: 
\vspace{-2mm}
\begin{equation}\label{Eq_disparity2xyz}
    \begin{aligned}
        z &= f \cdot b \cdot \frac{1}{\boldsymbol{D}(u,v)} , \quad 
        x &= z \cdot (u - c_x) \cdot \frac{1}{f}   , \quad
        y &= z \cdot (u - c_y) \cdot \frac{1}{f} ,
    \end{aligned}
\vspace{-2mm}
\end{equation}
where $c_x, c_y$ are the camera centroid, $f$ is focal length, and $b$ is the baseline length of the stereo camera, which can all be obtained by camera calibration. With Eq.(\ref{Eq_disparity2xyz}), the disparity vector $\boldsymbol{q} \in \mathbb{R}^3 = (u, v, D(u, v))^T$ is projected to a spacial point $\boldsymbol{p} \in \mathbb{R}^3 = (x,y,z)^T$. Given the color information from its corresponding RGB image $I_L$, we can obtain the point cloud representation of the surgical scene as $\boldsymbol{p} \in \mathbb{R}^6 = (x,y,z,r,g,b)^T$.
Then, we leverage renderer's shader to render and visualize the point cloud of the surgical scene. For correctly rendering the point cloud frame in the HMD view, it is important to perform coordinate transformations. In this regard, we propose the following scheme as:
\vspace{-2mm}
\begin{equation}
    \boldsymbol{\dot{p}_H} = \boldsymbol{K}_W^H \cdot \boldsymbol{K}_O^W \cdot \boldsymbol{K}_C^O \cdot \boldsymbol{\dot{p}}, \quad 
    \boldsymbol{K} = 
    \left(
        \begin{array}{cc}
             \boldsymbol{R} & \boldsymbol{t} \\
             \boldsymbol{0}^T & 1
        \end{array}
    \right), \quad
    \boldsymbol{\dot{p}} = (\boldsymbol{p}^T|1)^T,
\vspace{-2mm}
\end{equation}
where $\boldsymbol{p}_H$ and $\boldsymbol{p}$ are respectively the points in the HMD and camera coordinate frame, the dot notation means their corresponding homogenous vectors. $\boldsymbol{K}$ is transformation matrix
where $\boldsymbol{R}$ and $\boldsymbol{t}$ denote rotation and translation shift within the euclidean group $SE3 := \left\{ \boldsymbol{R},\boldsymbol{t} | \boldsymbol{R} \in \mathbb{SO}_3, \boldsymbol{t} \in \mathbb{R}^3 \right\}$. For $\boldsymbol{K}$, We use different subscripts to indicate different coordinate system, $\{O\}$ denotes the rendering centroid which is defined by box boundary of the first point cloud frame, $\{C\}$ is the camera coordinate frame, $\{W\}$ is the real world coordinate system, $\{H\}$ is the HMD coordinate frame. For instance, $\boldsymbol{K}_W^H$ indicates the transformation matrix from the real world to HMD coordinate system. In our case, $\boldsymbol{K}_C^O$ is fixed and $\boldsymbol{K}_W^H$ can be automatically obtained from HMD. Users can flexibly change the location and scale of the point cloud scene by $\boldsymbol{K}_O^W$ for interaction.

Importantly, we design all the mentor's interactions to be conducted purely by human hand which is highly user-friendly. Specifically, the cameras on HMD will capture multi-view images, compute the depth information, and reconstruct the environment. Then we employ built-in computer vision algorithms to analyze the spatial and pose information to recognize hand gestures, and locate the hand position w.r.t. the visualized virtual scene to achieve a hand-based interaction. For example, we design to raise the hand for waking up the setting menu, move finger in the air for pointing to certain critical anatomy in the scene. Moreover, we create assisting virtual objects (e.g., needle) with AR to facilitate necessary demonstration provided by the mentor. We design CAD models of the virtual objects and import them into HMD. With our system, the mentor can manipulate the virtual object purely using a hand, just like naturally moving an object in the real-world. For example, the mentor can manipulate a virtual needle with a pinch to rotate and move, and draw lines as instructions of suturing trajectories. These interactions can be recorded in real-time as the mentor's feedback.

\vspace{-2mm}
\subsection{Closed-loop Feedback for Guidance in Surgeon Console}
\vspace{-2mm}
To feedback the guidance information of the mentor, 
we record the hand interactions and virtual objects, and transmit them back to the operation side. Instead of directly sending rendered image with overlaid guidance information, we again design a light-weight data transfer approach to boost the efficiency. Specifically, we only send well-formatted location information to operation side, which can be the position of the virtual surgical tool, position of the draw trajectory, etc. The format of the feedback data is as $\{ m, i, \alpha_C, \beta_C, \gamma_C, x_C, y_C, z_C \}$, where $m$ is a flag to denote whether this information is available, $i$ is used to decide which kinds of information is contained (such as different kinds of virtual 
object), the $(\alpha_C, \beta_C, \gamma_C)$ means the yaw, pitch, and roll angle in the camera coordinate frame respectively, the $(x_C, y_C, z_C)$ represents the position of 
specified object. The feedback message is also sent to operating side using the TCP/IP Socket. After receiving the feedback information, operation side will decode the message and overlay related guidance information on the stereo video, and finally input to TilePro in the surgical robot console as augmented view to be perceived.

In this way, we build our system as an effective close-loop education paradigm. The mentor of expert surgeon can observe remote surgical procedures by simply wearing an AR glass. By manipulating virtual objects or indicating instructions with hands, the mentor can easily provide supportive advice to trainees in real-time. This new AR-based remote mentoring solution is distinct from traditional ones that use 2D video or haptic device. We have advantages of immersive visual experience, natural interaction, high-speed and low-cost, as well as great potential for richer information to be added on top of the augmented environment.




\vspace{-3mm}
\section{Experiments}

\vspace{-2mm}
\subsection{System Setup and Experiment Design}
\vspace{-2mm}

\begin{figure}[t] 
    \centering
    \vspace{-5mm}
\includegraphics[width=0.9\textwidth]{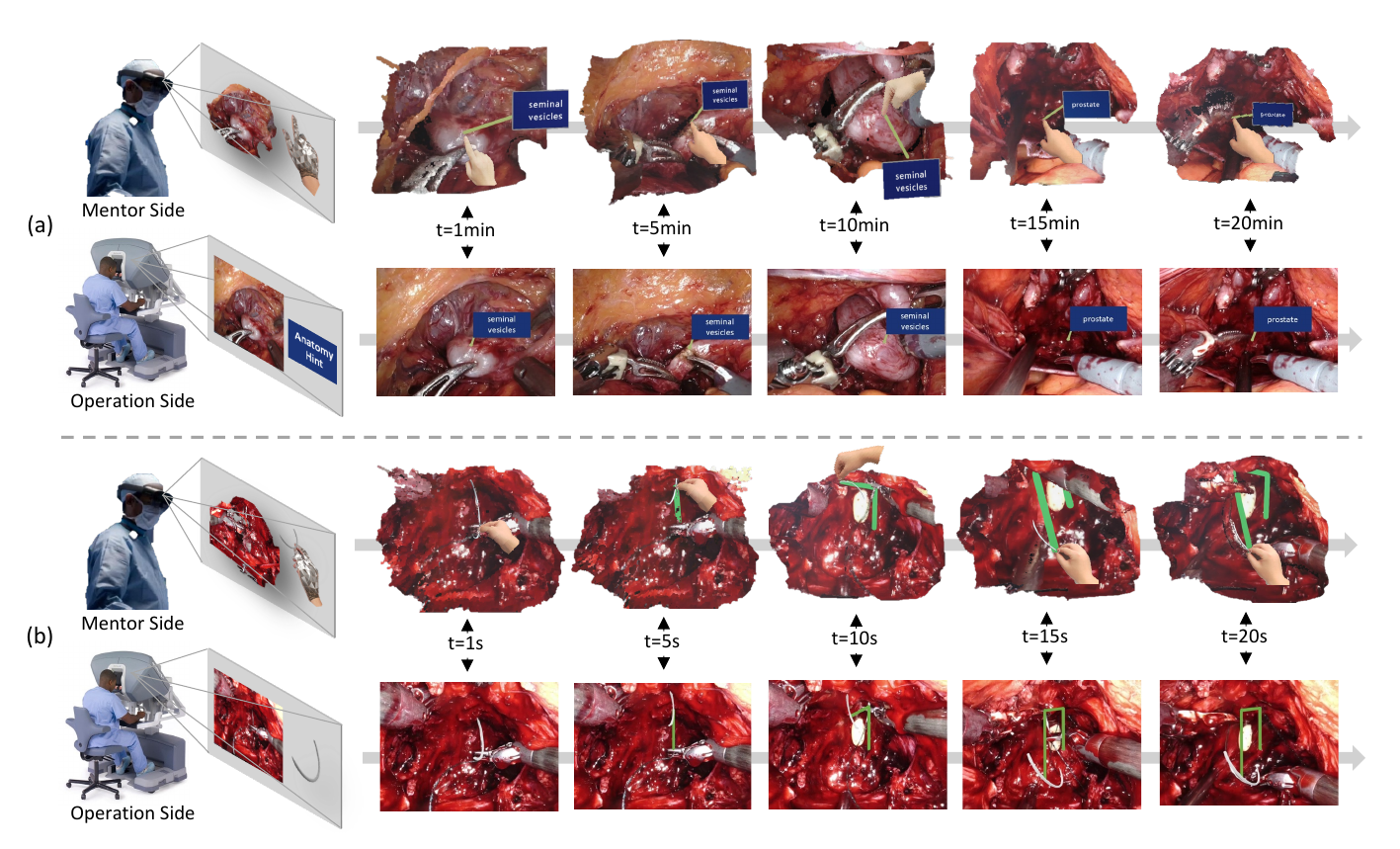}
\vspace{-6mm}
\caption{Case studies on real surgery data of prostatectomy. (a) Mentor uses the hand to point key anatomies in surgical scene which can be shown in the operation console for the trainee. (b) Mentor manipulates a virtual needle and draw suturing trajectories (green line) which can be demonstrated to the trainee. More details are in Sec.~\ref{Section_exp}.} 
\label{Fig_realsurgery}
\vspace{-3mm}
\end{figure}

For system setup, we use an in-house \emph{daVinci Si} surgical robot as the operation side, which is retired but generally similar to the robot used in clinic.
Its stereo video is streamed to a computer equipping NVIDIA TITAN RTX GPU for computing 3D representation of the surgical scene. 
The mentor side uses a \emph{Microsoft HoloLens 2}, the prevalent portable AR device, to visualize the 3D surgical scene and provide hand interaction with Unity toolkit.
We set up the wireless network using a low-cost WIFI router (TP-LINK AX3000) to connect the computer and HoloLens for remote surgical scene streaming and guidance feedback. 

For experiment design, we extensively validate our proposed system on both real surgery stereo videos and ex-vivo scenarios of common robotic training tasks.
For real surgery, we use two typical cases of DaVinci robotic surgery of prostatectomy procedure, with stereo videos captured in the resolution of $1280 \! \times \! 1024$ at $60fps$. To reduce computation cost of disparity map with HITNet~\cite{tankovich2021hitnet}, we first downsample the resolution to $640 \! \times \! 512$. We play the video in its original speed for simulation of the real clinical scenario with our system setup. The HoloLens can render the scene at the range of $25 \! \sim \! 30 fps$ for the mentor side.
For ex-vivo scenarios with \emph{daVinci Si}, we capture stereo videos with the same resolution and frame rate as the real surgery setting. We implement two scenarios, i.e., peg-transfer and suturing, which are common tasks for surgical skill training~\cite{joseph2010chopstick}. In peg-transfer, the trainee picks up the block from one peg and moves it to another target peg. In suturing, the trainee manipulates the needle to suture the phantom wound. We use these ex-vivo experiments for quantitative evaluations of the interaction accuracy and remote mentoring latency of our system.

\vspace{-2mm}
\subsection{Feasibility Study on Real Surgical Scenes}
\label{Section_exp}
\vspace{-1.5mm}
To demonstrate the application scenario of our remote mentoring system in the real clinical environment, we show two case studies with the prostatectomy procedure. As depicted in Fig.~\ref{Fig_realsurgery}(a), the surgeon is conducting the step of ``dissection of the seminal vesicles and rectum" in prostatectomy.
Inexperienced surgeons can meet difficulties to precisely identify the positions of key anatomies for dissection, due to fuzzy blood occlusion and poor lighting condition.
With our system, the expert can monitor the procedure for a minute-long duration, and point to the key anatomies in the air with hand from time to time. These hints are feedback to the operator's console for guidance.
Similarly, in Fig.~\ref{Fig_realsurgery}(b), the surgical step is ``van velthoven anastomosis and rocco stitch", in which suturing is demanding. The surgeon needs to incorporate as much of the muscular structural support behind the urethra as possible, therefore has to handle multiple suturing targets and complex needle trajectories which always require expertise. Using our system, the expert can provide help with a hand holding a virtual needle to indicate the next suturing target and demonstrate suturing trajectory.
Such feedback together with the virtual needle can again be transmitted back to operation console for helping trainees.
These results on real surgical data present the fidelity of streamed surgical scenes and flexibility of hand interactions, thus demonstrating promising feasibility for application in clinical practice.



\begin{table}[t]
\begin{center}
\caption{The accuracy results for two ex-vivo scenarios.}
\vspace{-2mm}
\label{acc_table}
\resizebox{0.6\textwidth}{!}
{
\begin{tabular}{c|cc}
\toprule
Scenario & ~Peg-transfer~  & ~Suturing~ \\
\hline
~Re-projection error (pixel)~       & 7.35 $\pm$ 2.1 & 5.12 $\pm$ 1.47  \\
\bottomrule
\end{tabular}
}
\end{center}
\vspace{-7mm}
\end{table}

\begin{figure}[t]
    \centering
\includegraphics[width=0.9\textwidth]{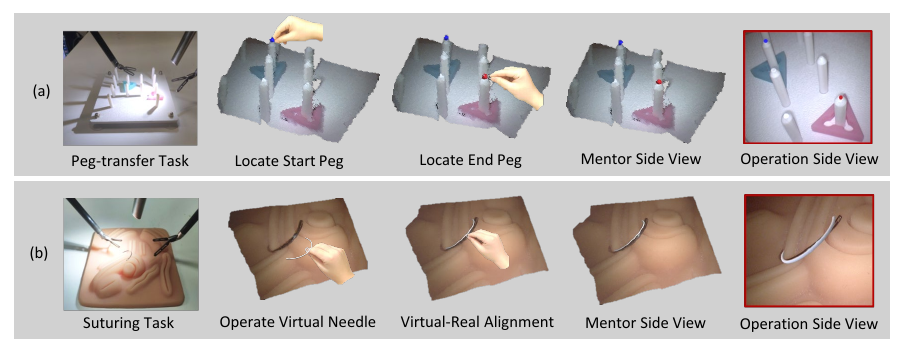}
\vspace{-5mm}
\caption{Two scenarios for testing the accuracy, with peg-transfer evaluating pointing accuracy and suturing evaluating the virtual needle interaction accuracy.}
\vspace{-6mm}
\label{Fig_scenario}
\end{figure}

\vspace{-3mm}
\subsection{Quantitative Evaluation on Ex-vivo Scenarios}
\vspace{-1mm}

\subsubsection{Accuracy evaluation.}

To evaluate how accurate the hand-based interactions can achieve in this system, we conduct following experiments. 
Specifically, we evaluate the pointing accuracy using peg-transfer scenario, where mentor points to the position of starting peg and ending peg with hand in HoloLens, and then feedback the position to surgeon's console in operation side, as shown in Fig.~\ref{Fig_scenario}(a). We first manually locate 2D location of the starting and ending peg tips as ground truth. Then we get the hand-indicated 3D position from HoloLens and re-project it to 2D location on console view. We then calculate the error between re-projected 2D location and ground truth. In this way, we can obtain the close-loop pointing error in operation side. We conduct 30 trials and average the final results. As shown in Tab.~\ref{acc_table}, the position error is $7.35$ pixel in average, which is hard to be perceived by human eyes and also within acceptable range compared 
with HoloLens calibration error~\cite{qian2017comprehensive}, thus precise enough for pointing. 

Moreover, we evaluate the interaction accuracy of hand manipulation of virtual-object using suturing scenario, where the mentor side is asked to move a virtual needle with hand and align it to the real needle seen in the HoloLens, as shown in Fig.~\ref{Fig_scenario}(b). Similarly, we manually locate 2D location of needle as ground truth, then record the ``aligned'' pose of virtual needle and re-project it to 2D console view as ``aligned" 2D location. 
We calculate the error between 'aligned' 2D location and ground truth. In this way, we can obtain the close-loop virtual-object based interaction error in operation side. We also conduct 30 trials and average the results. As shown in Tab.~\ref{acc_table}, 2D re-projection error is $5.12$ pixel in average, also consistent to HoloLens calibration error~\cite{qian2017comprehensive}, thus precise enough for virtual-object based interaction of surgical mentoring. 




\vspace{-4mm}
\subsubsection{Latency evaluation.}


\begin{table}[t]
\begin{center}
\caption{The update frame rate in HoloLens under different transmission resolutions.}
\vspace{-2mm}
\label{latency_table}
\resizebox{0.8\textwidth}{!}
{
\begin{tabular}{c|ccccc}
\toprule
Resolution & ~1280$\times$720~  & ~960$\times$540~  & ~ 640$\times$480~  & ~ 480$\times$360~  & ~ 320$\times$240~  \\
\hline
FPS (Hz)     & 10.43  & 21.35 & 32.68 & 41.13 & 53.42  \\
\bottomrule
\end{tabular}
}
\end{center}
\vspace{-6mm}
\end{table}

\begin{figure}[t]
    \centering
\includegraphics[width=0.88\textwidth]{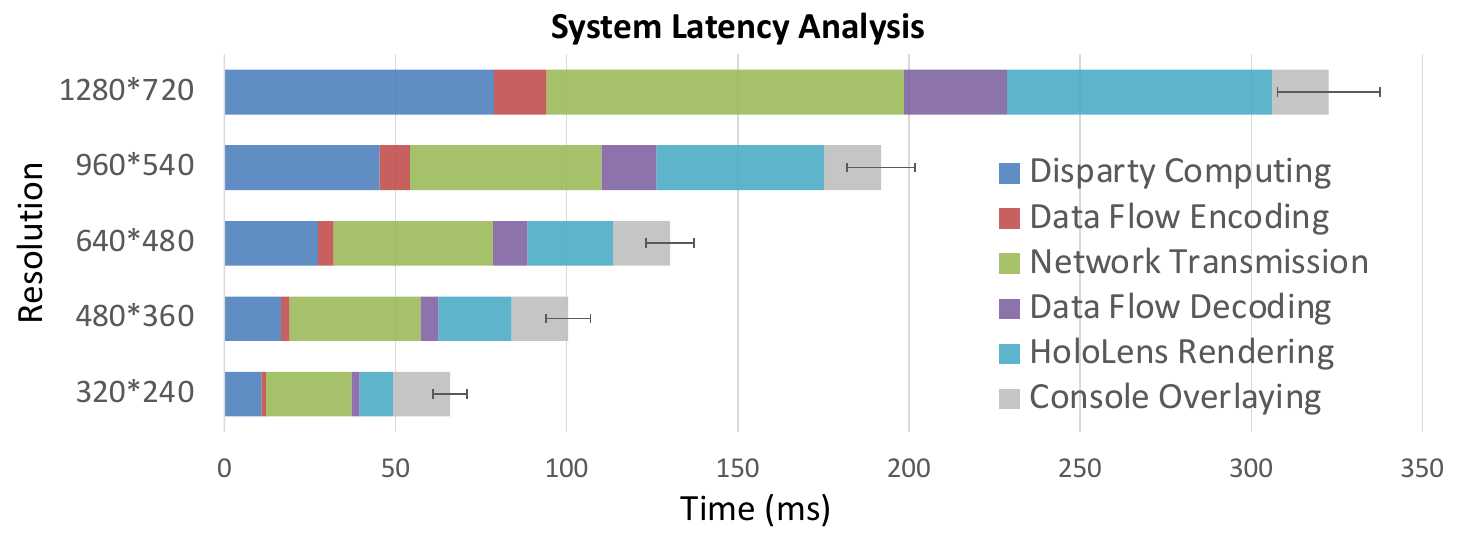}
\vspace{-4mm}
\caption{The latency analysis of each component in our proposed system.}
\label{Fig_latency}
\vspace{-6mm}
\end{figure}


To thoroughly evaluate the overall latency, we test our system under different common-used video resolutions from $320 \! \times \! 240$ to $1280 \! \times \! 720$. Specifically, Tab.~\ref{latency_table} reports the HoloLens update frame rate in mentor side, which can achieve satisfying real-time update speed ($32.68 fps$) with standard definition resolution ($640 \!\times \! 480$). Even when given high definition resolution ($1280 \!\times \!720$), the HoloLens can still achieve acceptable update rate ($10.43 fps$) to fulfill the low requirement of mentor side for basic visual-hand interaction. We also evaluate the end-to-end closed-loop latency for operation side and distinguish different components' latency, as shown in Fig~\ref{Fig_latency}. The total latency ranges from $65.9\pm5.0~ms$ to $322.6\pm15.1~ms$ with different resolution with even the highest latency is within the acceptable latency ($330~ms$) in terms of surgeon's perception of safety according to prior studies~\cite{lacy20195g,marescaux2001transatlantic}, thus feasible and efficient for remote mentoring. We further discuss the latency of each component. Relative constant time for console overlaying is observed, as GPU can overlay guidance parallelly. And the disparity computing and HoloLens rendering time decrease dramatically when given lower resolution, showing the high efficiency of our system.





\vspace{-3mm}
\section{Conclusion and Future Work}
\vspace{-3mm}
In conclusion, we present the first completely AR-based remote mentoring system with a head-mounted display for robotic surgical education.
Experiments and the achieved promising results have demonstrated its potential as a new reliable, flexible and low-cost solution for next-generation remote mentoring.
For future work, we will extend the system to allow cooperative two-hands interaction, in order to facilitate mentoring for more complex procedures. Moreover, we will conduct systematic user study with our surgeons, and they have already expressed interest and potential usefulness for their clinical practice.

%
%

\section{Acknowledgment}
This project was supported by CUHK Shun Hing Institute of Advanced Engineering (project MMT-p5-20), Hong Kong RGC TRS Project (No.T42- 409/18-R), Science, Technology and Innovation Commission of Shenzhen Municipality Project No. SGDX20220530111201008, and InnoHK Multi-Scale Medical Robotics Center.
%
%
%
\bibliographystyle{splncs04}
\bibliography{refs}

\begin{thebibliography}{10}
\providecommand{\url}[1]{\texttt{#1}}
\providecommand{\urlprefix}{URL }
\providecommand{\doi}[1]{https://doi.org/#1}

\bibitem{barmaki2019enhancement}
Barmaki, R., Yu, K., Pearlman, R., Shingles, R., Bork, F., Osgood, G.M., Navab,
  N.: Enhancement of anatomical education using augmented reality: An empirical
  study of body painting. Anatomical sciences education  \textbf{12}(6),
  599--609 (2019)

\bibitem{chen2021evolving}
Chen, I., Alan, H., Ghazi, A., Sridhar, A., Stoyanov, D., Slack, M., Kelly,
  J.D., Collins, J.W.: Evolving robotic surgery training and improving patient
  safety, with the integration of novel technologies. World Journal of Urology
  \textbf{39}(8),  2883--2893 (2021)

\bibitem{collins2021telementoring}
Collins, J.W., Ma, R., Beaulieu, Y., Hung, A.J.: Telementoring for minimally
  invasive surgery. In: Digital Surgery, pp. 361--378. Springer (2021)

\bibitem{d2021accelerating}
D'Ettorre, C., Mariani, A., Stilli, A., Valdastri, P., Deguet, A., Kazanzides,
  P., Taylor, R.H., Fischer, G.S., DiMaio, S.P., Menciassi, A., et~al.:
  Accelerating surgical robotics research: Reviewing 10 years of research with
  the dvrk. arXiv preprint arXiv:2104.09869  (2021)

\bibitem{erridge2019telementoring}
Erridge, S., Yeung, D.K., Patel, H.R., Purkayastha, S.: Telementoring of
  surgeons: a systematic review. Surgical innovation  \textbf{26}(1),  95--111
  (2019)

\bibitem{hinata2014novel}
Hinata, N., Miyake, H., Kurahashi, T., Ando, M., Furukawa, J., Ishimura, T.,
  Tanaka, K., Fujisawa, M.: Novel telementoring system for robot-assisted
  radical prostatectomy: impact on the learning curve. Urology  \textbf{83}(5),
   1088--1092 (2014)

\bibitem{huang2019telemedicine}
Huang, E.Y., Knight, S., Guetter, C.R., Davis, C.H., Moller, M., Slama, E.,
  Crandall, M.: Telemedicine and telementoring in the surgical specialties: a
  narrative review. The American Journal of Surgery  \textbf{218}(4),  760--766
  (2019)

\bibitem{jarc2017proctors}
Jarc, A.M., Stanley, A.A., Clifford, T., Gill, I.S., Hung, A.J.: Proctors
  exploit three-dimensional ghost tools during clinical-like training
  scenarios: a preliminary study. World journal of urology  \textbf{35}(6),
  957--965 (2017)

\bibitem{joseph2010chopstick}
Joseph, R.A., Goh, A.C., Cuevas, S.P., Donovan, M.A., Kauffman, M.G., Salas,
  N.A., Miles, B., Bass, B.L., Dunkin, B.J.: “chopstick” surgery: a novel
  technique improves surgeon performance and eliminates arm collision in
  robotic single-incision laparoscopic surgery. Surgical endoscopy
  \textbf{24}(6),  1331--1335 (2010)

\bibitem{kovoor2021validity}
Kovoor, J.G., Gupta, A.K., Gladman, M.A.: Validity and effectiveness of
  augmented reality in surgical education: A systematic review. Surgery  (2021)

\bibitem{lacy20195g}
Lacy, A., Bravo, R., Otero-Pi{\~n}eiro, A., Pena, R., De~Lacy, F., Menchaca,
  R., Balibrea, J.: 5g-assisted telementored surgery. British Journal of
  Surgery  \textbf{106}(12),  1576--1579 (2019)

\bibitem{marescaux2001transatlantic}
Marescaux, J., Leroy, J., Gagner, M., Rubino, F., Mutter, D., Vix, M., Butner,
  S.E., Smith, M.K.: Transatlantic robot-assisted telesurgery. Nature
  \textbf{413}(6854),  379--380 (2001)

\bibitem{mendez2005robotic}
Mendez, I., Hill, R., Clarke, D., Kolyvas, G., Walling, S.: Robotic
  long-distance telementoring in neurosurgery. Neurosurgery  \textbf{56}(3),
  434--440 (2005)

\bibitem{navab2021robotic}
Navab, N.: Robotic imaging, machine learning and augmented reality for computer
  assisted interventions. In: Medical Imaging 2021: Imaging Informatics for
  Healthcare, Research, and Applications. vol. 11601, p. 1160103. International
  Society for Optics and Photonics (2021)

\bibitem{qian2017comprehensive}
Qian, L., Azimi, E., Kazanzides, P., Navab, N.: Comprehensive tracker based
  display calibration for holographic optical see-through head-mounted display.
  arXiv preprint arXiv:1703.05834  (2017)

\bibitem{qian2019review}
Qian, L., Wu, J.Y., DiMaio, S.P., Navab, N., Kazanzides, P.: A review of
  augmented reality in robotic-assisted surgery. IEEE Transactions on Medical
  Robotics and Bionics  \textbf{2}(1),  1--16 (2019)

\bibitem{qian2019aramis}
Qian, L., Zhang, X., Deguet, A., Kazanzides, P.: Aramis: Augmented reality
  assistance for minimally invasive surgery using a head-mounted display. In:
  International Conference on Medical Image Computing and Computer-Assisted
  Intervention. pp. 74--82. Springer (2019)

\bibitem{shabir2021towards}
Shabir, D., Abdurahiman, N., Padhan, J., Trinh, M., Balakrishnan, S., Kurer,
  M., Ali, O., Al-Ansari, A., Yaacoub, E., Deng, Z., et~al.: Towards
  development of a tele-mentoring framework for minimally invasive surgeries.
  The International Journal of Medical Robotics and Computer Assisted Surgery
  \textbf{17}(5),  e2305 (2021)

\bibitem{tankovich2021hitnet}
Tankovich, V., Hane, C., Zhang, Y., Kowdle, A., Fanello, S., Bouaziz, S.:
  Hitnet: Hierarchical iterative tile refinement network for real-time stereo
  matching. In: Proceedings of the IEEE/CVF Conference on Computer Vision and
  Pattern Recognition. pp. 14362--14372 (2021)

\bibitem{yoon2018augmented}
Yoon, J.W., Chen, R.E., Kim, E.J., Akinduro, O.O., Kerezoudis, P., Han, P.K.,
  Si, P., Freeman, W.D., Diaz, R.J., Komotar, R.J., et~al.: Augmented reality
  for the surgeon: systematic review. The International Journal of Medical
  Robotics and Computer Assisted Surgery  \textbf{14}(4),  e1914 (2018)

\end{thebibliography}

\end{document}